# div2vec: Diversity-Emphasized Node Embedding

Jisu Jeong
Clova AI Research, NAVER Corp.
Seongnam, South Korea
jisu.jeong@navercorp.com

Jeong-Min Yun
WATCHA Inc.
Seoul, South Korea
matthew@watcha.com

Hongi Keam
WATCHA Inc.
Seoul, South Korea
paul@watcha.com

Young-Jin Park
Naver R&D Center, NAVER Corp.
Seoul, South Korea
young.j.park@navercorp.com

Zimin Park
WATCHA Inc.
Seoul, South Korea
holden@watcha.com

Junki Cho
WATCHA Inc.
Seoul, South Korea
leo@watcha.com

## ABSTRACT

Recently, the interest of graph representation learning has been rapidly increasing in recommender systems. However, most existing studies have focused on improving accuracy, but in real-world systems, the recommendation diversity should be considered as well to improve user experiences. In this paper, we propose the diversity-emphasized node embedding *div2vec*, which is a random walk-based unsupervised learning method like DeepWalk and *node2vec*. When generating random walks, DeepWalk and *node2vec* sample nodes of higher degree more and nodes of lower degree less. On the other hand, *div2vec* samples nodes with the probability inversely proportional to its degree so that every node can evenly belong to the collection of random walks. This strategy improves the diversity of recommendation models. Offline experiments on the MovieLens dataset showed that our new method improves the recommendation performance in terms of both accuracy and diversity. Moreover, we evaluated the proposed model on two real-world services, WATCHA and LINE Wallet Coupon, and observed the *div2vec* improves the recommendation quality by diversifying the system.

## 1 INTRODUCTION

Most recommender system studies have focused on finding users' immediate demands; they try to build models that maximize the click-through rate (CTR). The learned system suggests high-ranked items that users are likely to click in a myopic sense [6, 9, 30, 32]. Such recommendation strategies have successfully altered simple popularity-based or handmade lists, thus being widely adopted on many online platforms including Spotify [15], Netflix [18], and so on.

However, the previous approaches have a potentially severe drawback, a lack of diversity. For example, consider a user just watched Iron Man. Since a majority of people tend to watch other Marvel Cinematic Universe (MCU) films like Iron Man 2, Thor, and Marvel's The Avengers after watching Iron Man, the system would recommend such MCU films based on historical log data. While the approach may lead to CTR maximization, 1) can we say that users are satisfied with these apparent results? Or, 2) would a wider variety of recommendations achieve better user experience?

Recently, a method that addresses the first question is presented on Spotify [2]. This work categorized those who listen to very similar songs and different sets of entities as *specialists* and *generalists*, respectively. This work observed that generalists are much more satisfied than specialists based on long-term user metrics (i.e., the conversion to subscriptions and retention on the platform). Thus, even if some users are satisfied with the evident recommendations (clicked or played), this satisfaction does not imply the users continue to use the platform.

To answer the second question, we propose the diversity-emphasized node embedding *div2vec*. Recently, the number of studies on graph structure [8, 10, 17, 24, 26, 28] is increasing. Unfortunately, most of those studies have merely focused on the accuracy. DeepWalk and *node2vec* are the first and the most famous node embedding methods [8, 24]. DeepWalk, *node2vec*, and our method *div2vec* generates random walks first and then use the Skip-gram model [20] to compute embedding vectors of all nodes. When generating random walks, their methods choose nodes of high degree more. It makes sense because if a node had many neighbors in the past, it will have many neighbors in the future, too. However, it may be an obstacle for personalizing. Using our new method, all nodes are evenly distributed in the collection of random walks. Roughly speaking, the key idea is to choose a node with weight $\frac{1}{d}$ where $d$ is the degree of the node. Also, we propose a variant of *div2vec*, which we call *rooted div2vec*, obtained by changing the weight $\frac{1}{d}$ to $\frac{1}{\sqrt{d}}$ in order to balance accuracy and diversity. To the best of our knowledge, our approach is the first node embedding method focusing on diversity. Details are in Section 3.

We evaluate our new methods on a benchmark and two real-word datasets. As the benchmark, we verify the offline metrics on the MovieLens dataset. As a result, *div2vec* got higher scores in the diversity metrics, such as coverage, entropy-diversity, and average intra-list similarity, and lower scores in the accuracy metrics, such as AUC score, than DeepWalk and *node2vec*. Furthermore, its variant *rooted div2vec* had the highest AUC score and also the diversity scores of *rooted div2vec* are the best or the second-best.

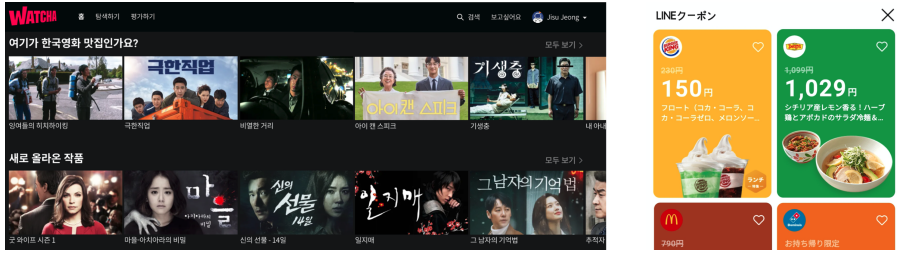

Figure 1: Screenshots of WATCHA and LINE Wallet Coupon.

We figure out that increasing diversity actually improves online performance. We test on two different live services, WATCHA and LINE Wallet Coupon. Screenshots of the services are in Figure 1. WATCHA is one of the famous OTT streaming services in South Korea. Like Netflix, users can watch movies and TV series using WATCHA. LINE is the most popular messenger in Japan, and LINE Wallet Coupon service provides various coupons, such as, pizza, coffee, shampoo, etc. In the above two different kinds of recommender systems, we used our diversity-emphasized node embedding and succeeded to enhance online performances. It is the biggest contribution of our work to prove that users in real world prefer a diverse and personalized recommendation.

The structure of the paper is as follows. In Section 2, we review random walk-based node embedding methods and the study on diversity problems. The proposed method will be described in Section 3. Section 4 and Section 5 reports the results of our experiments on offline tests and online tests, respectively. Section 6 concludes our research.

## 2 RELATED WORK
### 2.1 Random walk-based node embeddings

The famous *word2vec* method transforms words into embedding vectors such that similar words have similar embeddings. It uses a language model, called Skip-gram [20], that maximizes the co-occurrence probability among the words near the target word.

Inspired by *word2vec*, Perozzi et al. [24] introduced DeepWalk that transforms nodes in a graph into embedding vectors. A walk is a sequence of nodes in a graph such that two consecutive nodes are adjacent. A random walk is a walk such that the next node in the walk is chosen randomly from the neighbors of the current node. DeepWalk first samples a collection of random walks from the input graph where each node in random walks are chosen uniformly at random. Once a collection of random walks is generated, we treat nodes and random walks as words and sentences, respectively. Then by applying *word2vec* method, we can obtain an embedding vector of each node.

*node2vec* [8] is a generalization of DeepWalk. When nodes in random walks are chosen, *node2vec* uses two parameters $p$ and $q$. Suppose we have an incomplete random walk $v_1, v_2, \ldots, v_i$ and we will choose one node in the neighborhood $N(v_i)$ of $v_i$ to be $v_{i+1}$.

Here, for $x$ in $N(v_i)$, we set the weight $w(v_i, x)$ as follows:

$$w(v_i, x) = \begin{cases} \frac{1}{p} & \text{if } x = v_{i-1}, \\ 1 & \text{if } x \text{ is adjacent to } v_{i-1}, \\ \frac{1}{q} & \text{otherwise.} \end{cases}$$

Note that if a graph is bipartite, the second case does not appear. *node2vec* chooses $v_{i+1}$ at random with the weight $w(v_i, x)$.

The most advantage of graph representation learning or graph neural networks is that these models can access both local and higher-order neighborhood information. However, as the number of edges is usually too many, they may be inefficient. The random walk-based method solves this problem. Instead of considering all nodes and all edges, it only considers the nodes in the collection of random walks. Therefore, the way to generate random walks is important and it affects performance.

### 2.2 Diversity problems

The word "filter bubble" refers to a phenomenon in which the recommender system blocks providing various information and filters only information similar to the user's taste. In [3, 21, 22], they show the existence of the filter bubble in their recommender system. Some research [1, 27] claim that diversity is one of the essential components in the recommender system.

Some studies are proving that diversity increases the user's satisfaction. Spotify, one of the best music streaming services, observed that diverse consumption behaviors are highly associated with long-term metrics like conversion and retention [2, 13]. Also, Liu et al. [14] improve the user's preference by using neural graph filtering which learns diverse fashion collocation.

One may think that if a recommender system gains diversity, then it looses the accuracy. However, the following research succeeds in improving both. Adomavicius and Kwon [1] applied a ranking technique to original collaborative filtering in order to increase diversity without decreasing the accuracy. Zheng et al. [31] proposed a Deep Q-Learning based reinforcement learning framework for news recommendation. Their model improves both click-through rate and intra-list similarity.

## 3 PROPOSED METHOD
### 3.1 Motivation

In the framework of DeepWalk and *node2vec*, the model first generates a collection of random walks, and then runs the famous *word2vec* algorithm to obtain embedding vectors. In their way, nodes of high degree should be contained more than nodes of low degree in the collection of random walks because, roughly speaking, if a node $v$ has $d$ neighbors, then there are $d$ chances that $v$ can belongs to the collection of random walks. Figure 2a and Figure 2b represent this phenomenon. The $x$-axis are the nodes sorted by the degree and the blue line means the degree of nodes. So, the blue line is always increasing and it means that nodes of higher degrees are on the right side in each figure. Orange bars mean the frequencies of nodes in the collection of random walks. It is easy to observe that nodes of high degree appear extremely more than that of low degree.

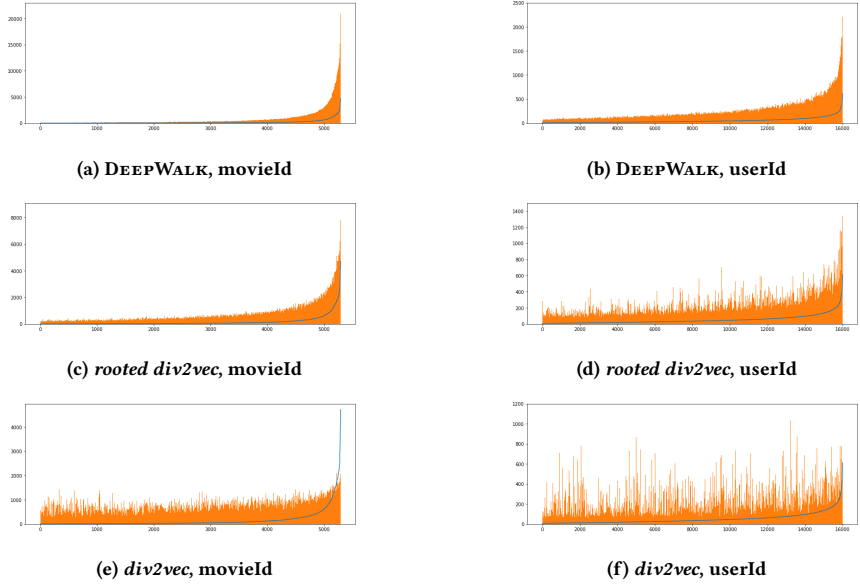

Figure 2: The $x$-axis are nodes and the blue line means the degree of nodes. The $y$-axis denotes the frequency of nodes in the collection of random walks.

As the collection of random walks are the training set of the skim-gram model, DeepWalk and *node2vec* might be trained with a bias to nodes of high degree. It may not be a trouble for solving problems focused on the accuracy. For example, in link prediction, high-degree nodes in the original graph might have a higher probability of being linked with other nodes than low-degree nodes. In terms of movies, if the movie is popular, then many people will like this movie. However, it must be a problem when we want to focus on personalization and diversity. Unpopular movies might not be trained enough so that they are not well-represented. So, even if a person actually prefers some unpopular movie to some popular movie, the recommender system tends to recommend the popular movie for safe.

Motivated by Figure 3, which shows the difference between reality and equity, we decided to consider the degree of the next candidate nodes inversely. The main idea is 'Low degree, choose more.'. We propose a simple but creative method, which will be formally described in the next subsection, which gives Figure 2e and Figure 2f. Compare to Figure 2a and Figure 2b, the nodes in

Figure 2e and Figure 2f are evenly distributed regardless of their degree.

### 3.2 div2vec

Now, we introduce the diversity-emphasized node embedding method. Similarly to DeepWalk and *node2vec*, we first produce a lot of random walks and train skip-gram model. We apply an easy but bright idea to generate random walks so that our model can capture the diversity of the nodes in their embedding vectors.

Suppose a node $v$ is the last node in an incomplete random walk and we are going to choose the next node among the neighbors of $v$.

- DeepWalk picks the next node in $N(v)$ at random with the same probability.
- In *node2vec*, if $w$ is the node added to the random walk just before $v$, then there are three types of probability depend on whether a node $u \in N(v)$ is adjacent with $w$ or not, or $u = w$.
- Our method will choose the next node according to the degree of neighbors.

Formally, our method chooses the next node $u \in N(v)$ with the probability

$$\frac{f(\deg(u))}{\sum_{w \in N(v)} f(\deg(w))}$$

for some function $f$. For example, when $f(x) = 1/x$, if $x$ has two neighbors $y$ and $z$ whose degree is 10 and 90 respectively, then $y$ is chosen with probability $(1/10)/(1/10 + 1/90) = 0.9$ and $z$ is chosen with probability 0.1. That is, since the degree of $y$ is smaller than the degree of $z$, the probability that $y$ is chosen is larger than the probability that $z$ is chosen. In Section 4, we set $f$ to the inverse of the identity function $f(x) = 1/x$ and the inverse of the square root function $f(x) = 1/\sqrt{x}$. We call this method *div2vec* when $f(x) = 1/x$ and *rooted div2vec* when $f(x) = 1/\sqrt{x}$.

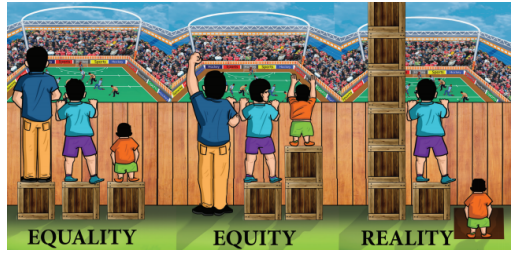

**Figure 3: There are three people of different heights. The concept of equality is to give the same number of boxes to all. However, in reality, the rich get richer and the poor get poorer. As a perspective of equity, the small person get more boxes than the tall person.**[1]

Intuitively, DeepWalk chooses the next node without considering the past or the future nodes, *node2vec* chooses the next node according to the past node, and *div2vec* chooses the next node with respect to the future node. Note that it is possible to combine *node2vec* and *div2vec* by first dividing into three types and then consider the degree of neighbors. Since there are too many hyperparameters to control, we remain it to the next work.

Figure 2 is the result for generating random walks with several methods. The detail for the dataset is in Subsection 4.1. If we use DeepWalk, then Figure 2a and Figure 2b show that high-degree nodes appears extremely much more than low-degree nodes. The problem is that, if the result is too skew, then the skip-gram model might not train some part of data well. For example, a popular movie will appear many times in the collection of random walks and then the model should overfit to the popular movie. On the other hands, an unpopular movie will appears only few times in the collection of random walks and then the model should underfit to the unpopular movie.

This problem is solvable by using our method. Using *div2vec*, we can have the nodes evenly in the collection of random walks. Figure 2e and Figure 2f show that our method solves this problem. The nodes are chosen equally regardless of the degree of nodes. Normally, popular movies are consumed more than unpopular movies. So *div2vec* may decrease the accuracy. Our experiments prove that even if we emphasize the diversity, the accuracy decrease very little. Furthermore, we suggest the variant *rooted div2vec*. Figure 2c and Figure 2d can be treated as the combination of DeepWalk and *div2vec*. In Subsection 4.4, our experiments show that compare to DeepWalk and *node2vec*, *rooted div2vec* records better scores for every metric.

## 4 OFFLINE EXPERIMENTS
### 4.1 DataSets
We used the famous MovieLens dataset [11] for an offline test. We used MovieLens-25M because it is the newest data and we only used the recent 5 years in the dataset. Rating data is made on 10 steps, but we need binary data, which means watched/not or satisfied/unsatisfied, in order to train a model and compute AUC score. We set more than 4 stars to be positive and less than 3 stars to be negative. To prevent noises, we remove both the movies having less than 10 records and the users having less than 10 or more than 1000 records. At last, there are 2,009,593 records with 16,002 users and 5,298 movies. For the test set, 20% of the data are used.

To compute intra-list similarity, which will be described in Subsection 4.3, we use 'Tag Genome' [29] from MovieLens-25M. It contains data in 'movieId, tagId, relevance' format for every pair of movies and tags. Relevance values are real numbers between 0 and 1. So, it can be treated as a dense matrix and one row that represents one movie means a vector containing tag information.

### 4.2 Experiment settings
In movie recommender systems, a model recommends a list of movies to each user. In other words, a model needs to find out which movies will be connected with an individual user. It means that our task is a link prediction. However, the methods we discussed so far are only compute node embeddings. That is, we have an embedding vector for movies and users but not for their interactions. Grover and Leskovec [8] introduced four operators to obtain edge embeddings from node embeddings as follows. Let $u$ and $v$ be two nodes and $emb(u)$ and $emb(v)$ be their embedding vectors.

(1) Average: $\frac{emb(u)+emb(v)}{2}$
(2) Hadamard: $emb(u) * emb(v)$ (element-wise product)
(3) Weighted-L1: $|emb(u) - emb(v)|$
(4) Weighted-L2: $|emb(u) - emb(v)|^2$

For each edge, we obtain 64-dim vector from the graph induced by positive edges and 64-dim vector from the graph induced by negative edges. And then we concatenate the positive edge embedding vector and the negative edge embedding vector to represent the edge embedding vector.

To avoid disrupting the performance of a prediction model, we use simple deep neural network with one hidden layer of size 128.

### 4.3 Evaluation metrics
For each embedding and each operator, we compute four metrics, one for accuracy and the others for diversity. The larger score means the better performance.

---
[1]This figure is from http://www.brainkart.com/article/Equality_34271/.

| Method | AUC | CO(1) | ED(1) | CO(10) | ED(10) | ILS(10) | CO(50) | ED(50) | ILS(50) |
|---|---|---|---|---|---|---|---|---|---|
| DeepWalk | 0.874204 | 1035 | 4.882882 | 2944 | 6.051621 | 0.673236 | 4510 | 6.865239 | 0.670873 |
| n2v-(1,2) | 0.868537 | 1125 | 5.387989 | 2998 | 6.223997 | 0.685589 | 4483 | 6.877584 | 0.680990 |
| n2v-(2,1) | 0.864024 | 958 | 5.064918 | 2577 | 6.034628 | 0.682573 | 4081 | 6.773981 | 0.674534 |
| div2vec | 0.851322 | **2793** | **6.859013** | **4717** | 7.308675 | **0.706817** | 5243 | 7.614828 | 0.700030 |
| rooted div2vec | **0.888123** | 2332 | 6.713877 | 4500 | 7.275315 | 0.705358 | **5207** | **7.614992** | **0.700071** |

(a) The results with the operator Weighted-L1.

| Method | AUC | CO(1) | ED(1) | CO(10) | ED(10) | ILS(10) | CO(50) | ED(50) | ILS(50) |
|---|---|---|---|---|---|---|---|---|---|
| DeepWalk | 0.878293 | 1131 | 5.231090 | 3016 | 6.170295 | 0.665313 | 4453 | 6.837852 | 0.666019 |
| n2v-(1,2) | 0.871771 | 1302 | 5.504011 | 3261 | 6.323582 | 0.672085 | 4598 | 6.908540 | 0.671838 |
| n2v-(2,1) | 0.867549 | 973 | 4.865928 | 2725 | 5.942549 | 0.669559 | 4135 | 6.731078 | 0.667980 |
| div2vec | 0.853404 | **2435** | **6.544910** | **4608** | 7.196730 | **0.704196** | 5213 | 7.555974 | **0.700267** |
| rooted div2vec | **0.889674** | 2168 | 6.457435 | 4597 | **7.236516** | 0.700865 | **5233** | **7.612159** | 0.698983 |

(b) The results with the operator Weighted-L2.

| Method | AUC | CO(1) | ED(1) | CO(10) | ED(10) | ILS(10) | CO(50) | ED(50) | ILS(50) |
|---|---|---|---|---|---|---|---|---|---|
| DeepWalk | 0.862314 | 1331 | 6.088063 | 3156 | 6.856834 | 0.671195 | 4642 | 7.420330 | 0.667491 |
| n2v-(1,2) | 0.866459 | 1593 | 6.217412 | 3570 | 6.988375 | **0.684405** | 4740 | 7.478146 | **0.680929** |
| n2v-(2,1) | 0.863299 | 1587 | 6.315792 | 3371 | 7.053128 | 0.670602 | 4645 | 7.516015 | 0.669998 |
| div2vec | 0.837683 | **2605** | 7.092134 | **4623** | 7.711741 | 0.679191 | 5200 | 8.033483 | 0.673614 |
| rooted div2vec | **0.870787** | 2565 | **7.177497** | 4573 | **7.775564** | 0.670754 | **5254** | **8.103879** | 0.666571 |

(c) The results with the operator Hadamard.

| Method | AUC | CO(1) | ED(1) | CO(10) | ED(10) | ILS(10) | CO(50) | ED(50) | ILS(50) |
|---|---|---|---|---|---|---|---|---|---|
| DeepWalk | 0.903541 | 440 | 3.419942 | 1442 | 4.940299 | 0.709170 | 2872 | 5.988137 | 0.706576 |
| n2v-(1,2) | 0.907638 | 617 | 4.031595 | 1957 | 5.464436 | 0.730591 | 3481 | 6.403608 | 0.717133 |
| n2v-(2,1) | 0.909999 | 686 | 4.259565 | 2083 | 5.700572 | **0.745205** | 3540 | 6.471546 | 0.724843 |
| div2vec | 0.894085 | **882** | **4.730973** | **2575** | **6.130730** | 0.725977 | 4039 | **6.972115** | 0.715884 |
| rooted div2vec | **0.913342** | 831 | 4.695825 | 2481 | 6.083932 | 0.742770 | **4121** | 6.929550 | **0.727009** |

(d) The results with the operator Average.

Table 1: The results on an offline test.

**AUC SCORE** *AUC score* is area under the Receiver Operating Characteristic curve, which is plotting True Positive Rate (TPR) against False Positive Rate (FPR) at various thresholds. TPR and FPR are defined as follow:

$$\text{TPR} = \frac{(the\ number\ of\ true\ positive)}{(the\ number\ of\ true\ positive) + (the\ number\ of\ false\ negative)}$$

$$\text{FPR} = \frac{(the\ number\ of\ false\ positive)}{(the\ number\ of\ false\ positive) + (the\ number\ of\ true\ negative)}$$

AUC score is in range 0 to 1. As close to 1, it gives better evaluation and is close to perfect prediction. AUC score is useful evaluation metric because of scale invariant and classification threshold invariant to compare multiple prediction model.

**COVERAGE** *Coverage* is how many items appear in the recommended result. Formally, we can define as

$$\text{coverage}(M) = \left| \bigcup_u R_{M,k}(u) \right|$$

where $M$ is a model, $R_{M,k}$ is a set of top-$k$ recommended items for a user $u$ by $M$. Many papers [1, 7, 12, 16] discuss the importance of the coverage. If the coverage of the model is large, then the model recommends a broad range of items, and it implicitly means that users can have difference items. Furthermore, if the accuracy is competitive, then we may say that the model is good at personalization.

**ENTROPY-DIVERSITY** Adomavicius and Kwon [1] introduced the entropy-based diversity measure *Entropy-Diversity*. Let $U$ be the set of all users, and $rec_{M,k}(i)$ be the number of users $u$ such that $i \in R_{M,k}(u)$ for a model $M$, an integer $k$, and an item $i$. Then

$$\text{ENTROPY-DIVERSITY}(M) = -\sum_i \left( \frac{rec(i)}{k \times |U|} \right) \ln \left( \frac{rec(i)}{k \times |U|} \right).$$

Note that we can say that if ENTROPY-DIVERSITY($M_1$) < ENTROPY-DIVERSITY($M_2$), then $M_2$ recommends more diverse items than $M_1$. Here is an example. For an item set $I = \{item1, item2, \ldots, item9\}$ and a user set $U = \{user1, user2, user3\}$, suppose a model $M_1$ gives $R_{M_1,3}(u) = \{item1, item2, item3\}$ for every user $u$, and a model $M_2$ gives $R_{M_2,3}(user1) = \{item1, item2, item3\}$, $R_{M_2,3}(user2) = \{item4, item5, item6\}$, $R_{M_2,3}(user3) = \{item7, item8, item9\}$. Then ENTROPY-DIVERSITY($M_1$) $= -(3/9)\ln(3/9) \times 3 + 0 \times 6 = \ln 3$ and ENTROPY-DIVERSITY($M_2$) $= -(1/9)\ln(1/9) \times 9 = \ln 9$.

**AVERAGE INTRA-LIST SIMILARITY** From the recommendation model, every user will receive a list of items. *Intra-List Similarity (ILS)* measures how dissimilar or similar items in the list are.

| Week | clicks | plays | watch time |
|---|---|---|---|
| the first week | 66.19 | 62.07 | 3.58 |
| the second week | 39.69 | 28.52 | 4.19 |

Table 2: Percentage of improvements (%) of *div2vec* over *node2vec*.

Formally,
$$\text{ILS}(L) = \frac{\sum_{i \in L} \sum_{j \in L, i \neq j} \text{SIM}(i,j)}{|L|(|L|-1)/2}$$

where $L$ is the recommended item list and $\text{SIM}(i,j)$ is the similarity measure between the tag-genome vectors of $i$ and $j$, which are given from the MovieLens dataset [11, 29]. We set $\text{SIM}(i,j) = 1 - \frac{v_i \cdot v_j}{||v_i|| \cdot ||v_j||}$ where $v_i$ and $v_j$ are corresponding tag-genome vectors of $i$ and $j$. By definition, if the value is small, then items in the list are similar. Otherwise, they are dissimilar. Note that Bradly and Smyth [4], and Meymandpour and Davis [19] use the same definition in terms of 'diversity'. For every user $u$, we compute $ILS(R_{M,k}(u))$ and their average, which we call *Average Intra-List Similarity* in order to measure how diverse a model is.

### 4.4 Results on an offline test

Table 1 summarizes the results of our offline experiments on the MovieLens dataset. We did many experiments under various conditions.

- five methods: DeepWalk [24], node2vec [8] with different hyperparameters, *div2vec* and its variant *rooted div2vec*
- four operators: Weighted-L1, Weighted-L2, Hadamard, Average
- four metrics: AUC score, coverage, entropy-diversity, and average intra-list similarity
- three sizes of recommendation lists: 1, 10, 50

AUC means AUC score. $CO(k)$, $ED(k)$, $ILS(k)$ means coverage, entropy-diversity, average intra-list similarity of top $k$ recommended items, respectively. $n2v$-$(p,q)$ is *node2vec* with hyperparameter $p, q$.

Our proposed methods *div2vec* and *rooted div2vec* record the highest scores on all metrics in Table 1a and Table 1b In Table 1c and Table 1d, the average intra-list similarity is not the best but the second with tiny gaps. Overall, it is easy to see that *div2vec* and *rooted div2vec* got better scores than DeepWalk and *node2vec* in diversity metrics. Furthermore, *rooted div2vec* got the best scores in the accuracy metric. Thus, we can conclude that our proposed methods increase the diversity of recommender systems.

## 5 LIVE EXPERIMENTS

### 5.1 Video Recommendation

In the previous experiments, we verified that our methods, *div2vec* and *rooted div2vec*, clearly increase the diversity of recommended results. The remaining job is to prove that *div2vec* actually increases user satisfaction in real-world recommender systems. To show this, we conduct an A/B test in the commercial video streaming service, WATCHA, and measure and compare various user activity statistics that are related to user satisfaction.

We collected four months watch-complete logs starting from January 2020, here watch-complete means user completing a video. We filter-out users who do not have watch-complete logs last few days, also filter-out extreme case users (too many or too few logs); results in 21,620 users. Two methods, *node2vec* and *div2vec*, were trained with these filtered logs. For *node2vec*, we set the parameters $p = q = 1$.

An A/B test had been conducted two weeks, where 21,620 users were randomly and evenly partitioned into two groups and each group received either *node2vec* or *div2vec* recommending results. In more detail, WATCHA has list-wise recommendation home whose list consists of several videos, and our list inserted into the fifth row. To make the list, we sorted all available videos by the final scores and pick top $k$ of them ($k$ varies with devices), and also apply random shuffling of top $3k$ videos to avoid always the same recommendation.

We compare clicks and plays of *node2vec* and *div2vec* list by the week. (The first two columns in Table 2) In the first week, *div2vec* list received more than 60% more actions than *node2vec* list in both clicks and plays; 39.69% more clicks and 28.52% more plays at the second week. As we can see that *div2vec* beats *node2vec* with clicks and plays by a significant margin.

Someone may argue that the above improvement does not improve actual user satisfaction; if users who received the *node2vec* list are satisfied with other recommended list. To see this actually happens, we compare total watch time of each group during the test. (The last column in Table 2) In the first week, *div2vec* achieved 3.58% more watch time than *node2vec*, and 4.19% in the second week. Let me note that in watch time comparison, even 1% improvement is hard to achieve [5], thus our improvement is quite impressive results.

### 5.2 Coupon Recommendation

To further demonstrate the effectiveness of using the *div2vec* embedding in other real-world recommender systems, we run an A/B test in the LINE Wallet Coupon service and evaluate the online performance for two weeks in the spring of 2020. The system consists of over six million users and over five hundred unique items. We constructed the user-item bipartite graph by defining each user and item as an individual node and connecting nodes that are interacted each other. Using the graph, we obtained *div2vec* embedding vectors for each nodes. In this experiment, we compared the number of unique users clicked (Click UU) and click-through rate (CTR) of the neural-network based recommendation model[2] using the precomputed *div2vec* embedding vectors as additional features to the model that did not. As side information, the paper utilized gender, age, mobile OS type, and interest information for users, while brand, discount information, text, and image features for items. The online experiment results show that the overall Click UU and CTR have been improved by 6.55% and 2.90%, respectively. The relative performance for two weeks is illustrated in Figure 4 by date. By applying the *div2vec* feature, a larger number of users get interested in the recommended coupon list and the ratio that the user reacts to the exposed item increases, significantly. Considering that the online tests were conducted for a relatively long period, we conclude that the diversified recommendation based on the proposed method has led to positive consequences in user experience rather than to attract curiosity from users temporarily.

---
[2]The details in model architecture for the LINE Wallet Coupon recommender system are presented in [23, 25].

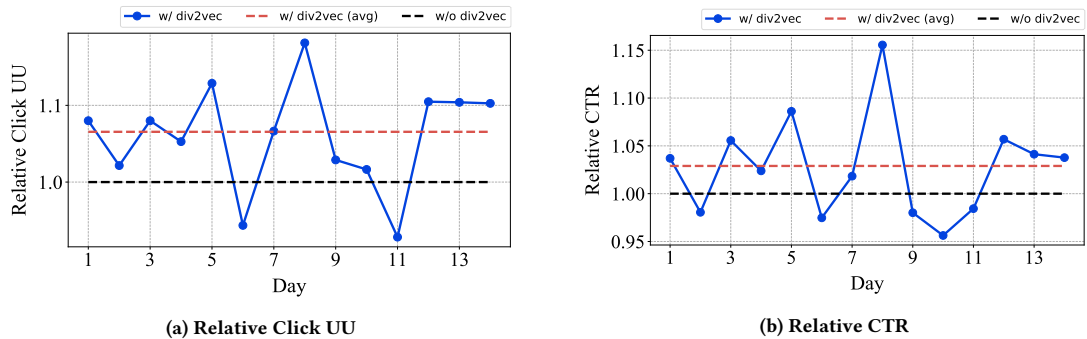

(a) Relative Click UU

(b) Relative CTR

Figure 4: Relative performance of the model applying *div2vec* feature to the existing model in LINE Wallet Coupon service by date.

## 6 CONCLUSION

We have introduced the diversity-emphasized node embedding *div2vec*. Several experiments showed the importance of our method. Compared to DeepWalk and *node2vec*, the recommendation model based on *div2vec* increased the diversity metrics like coverage, entropy-diversity, average intra-list similarity in the MovieLens dataset. The main contribution of this paper is that we verified that users satisfy with the recommendation model using *div2vec* in two different live services.

We remark that as *div2vec* is an unsupervised learning method like *word2vec*, it can be easily combined with other studies and services, and it is possible to improve their performance. Also, by changing the function $f$, the distribution of nodes in the collection of random walks can be adjusted to each domain.

## ACKNOWLEDGMENTS

Special thanks to those who lent their insight and technical support for this work, including Jaehun Kim, Taehyun Lee, Kyung-Min Kim, and Jung-Woo Ha.